\documentclass[11pt]{article}

\usepackage[final]{acl}

\usepackage{times}
\usepackage{latexsym}

\usepackage[T1]{fontenc}

\usepackage[utf8]{inputenc}

\usepackage{microtype}

\usepackage{inconsolata}

\usepackage{graphicx}
\usepackage{booktabs}
\usepackage{multirow}
\usepackage{ragged2e}
\usepackage{subcaption}
\usepackage{amsmath}
\usepackage{amsfonts}
\usepackage{svg}
\usepackage{enumitem}

\title{On the Limits of Steering Vectors for Preference-Aligned Generation}

\author{Melanie Subbiah$^*$ \and Zara Hall$^*$ \and Kathleen McKeown\\
        Department of Computer Science, Columbia University, USA\\
         \small{
   \textbf{Correspondence:} \href{mailto:m.subbiah@columbia.edu}{m.subbiah@columbia.edu}
 }}

\begin{document}
\maketitle
\begingroup
\renewcommand\thefootnote{}\footnotetext{$^*$ indicates equal contribution}
\endgroup
\begin{abstract}
Steering vectors have emerged as a promising approach to controlled text generation, offering interpretable, training-free mechanisms for shaping model outputs. However, their practical generality remains poorly understood. We study the limits of steering vector generalization along three dimensions: trait expressibility, task transfer, and multi-trait composition. Using the PLUME writing personalization benchmark, we extract steering vectors for a range of preferences and evaluate them on summarization and email-writing tasks across two open-source models (Qwen2.5-7B-Instruct and Llama3.1-8B-Instruct). We find that steering effectiveness varies substantially across traits. We further show that steering effectiveness can degrade when vectors extracted from positive and negative style examples are transferred
to downstream writing personalization tasks. Finally, we compare common methods for composing multiple steering vectors and find that all methods suffer significant drops in trait expression as more vectors are added, with a tradeoff between coherence and expressibility that requires per-setting hyperparameter tuning. Taken together, our results suggest that steering vectors face meaningful limits as a general-purpose tool for preference alignment.
\end{abstract}

\section{Introduction}

Steering vectors are directions in the model's activation space, which have been shown to be effective in both modeling and controlling model behavior \citep{chen2025personavectorsmonitoringcontrolling}. Steering vectors have several interesting advantages. While RL methods can update model weights to align them with specific preferences \citep{DBLP:journals/corr/abs-1909-08593, wei2022finetuned}, they require large datasets and costly fine-tuning. Steering vectors can be extracted from examples alone and added to the model at inference time, making them lightweight, flexible, and effective with just hundreds of synthetically generated examples \citep{turner2024steeringlanguagemodelsactivation,zou2025representationengineeringtopdownapproach}. Prompting can also be a lightweight method for adaptation, but LLMs can get ``lost" across long conversations and suffer from model drift  \citep{liu-etal-2024-lost, li2024measuring,choi2025examiningidentitydriftconversations}. Additionally, clever system or user prompts can be used to ``jailbreak'' models into problematic modes \citep{shen2024donowcharacterizingevaluating,liu2024autodan}. Because activation steering directly modifies internal representations at inference time, it has been explored as an alternative control mechanism to prompting \citep{turner2024steeringlanguagemodelsactivation,zou2025representationengineeringtopdownapproach,rimsky-etal-2024-steering}.

\begin{figure}
    \centering
    \includegraphics[width=\columnwidth]{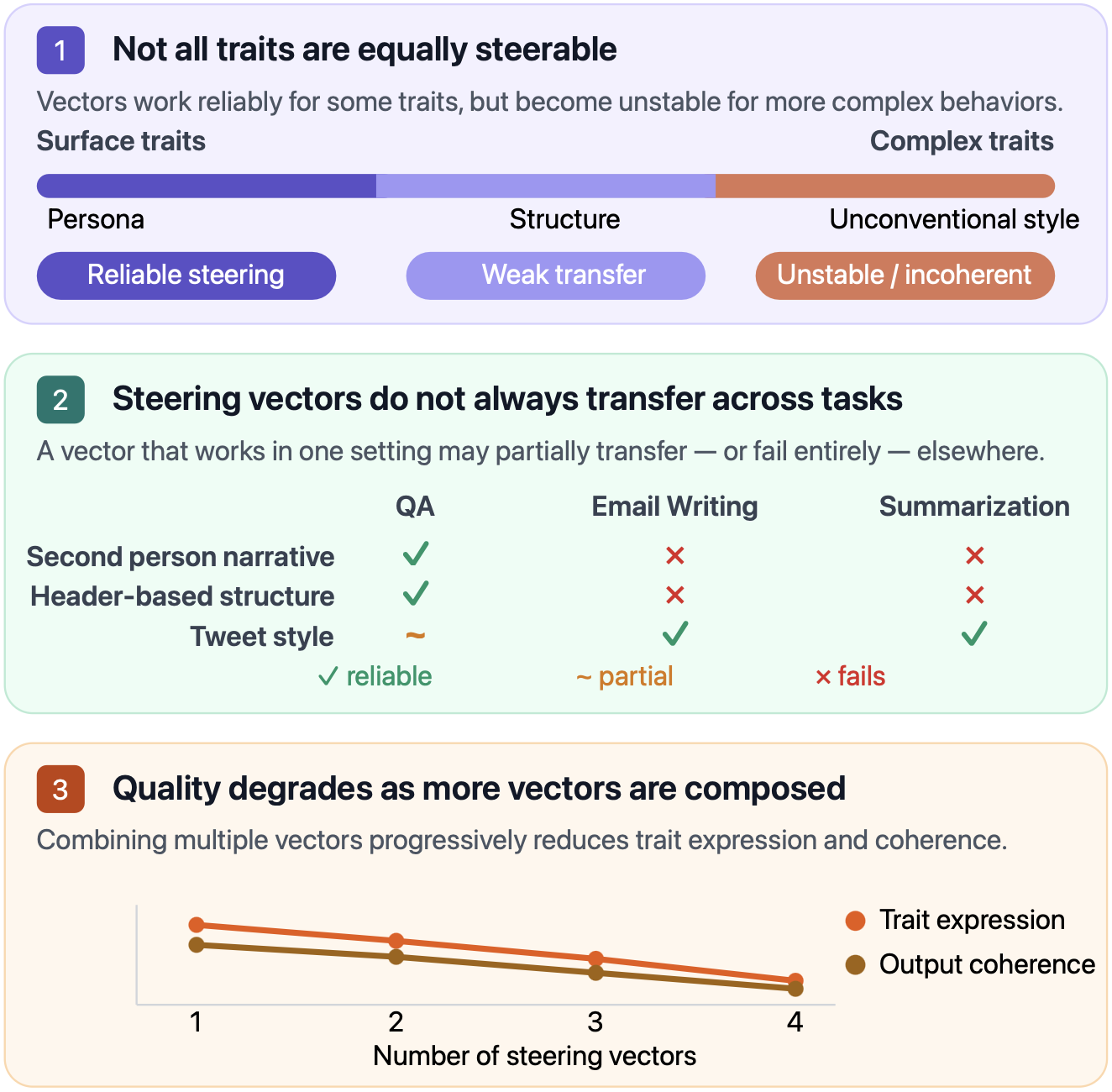}
    \caption{We systematically study the limitations of steering vectors when scaled across trait types, tasks, and number of vectors.}
    \label{fig:placeholder}
\end{figure}

In addition to controllable generation benefits, steering vectors provide interpretable insight into how the model is processing the input \citep{chen2025personavectorsmonitoringcontrolling}. When a certain input highly activates the direction in activation space corresponding to a given steering vector, it indicates the model is recognizing that vector's concept in the input \citep{zou2025representationengineeringtopdownapproach,chen2025personavectorsmonitoringcontrolling}. Examining how model outputs change when steered with a given steering vector also gives insight into how a model views a certain concept or persona. For example, steering in an \textit{evil} direction might make a model introduce more hallucinations into a summary. 

For all of these reasons, steering vectors have become a popular line of research in interpretability and alignment. However, most studies focus primarily on only a handful of traits to steer for, such as being \textit{helpful}, and generally only apply one steering vector at inference time \citep{turner2024steeringlanguagemodelsactivation,li2023inferencetime,rimsky-etal-2024-steering}. In reality though, for steering vectors to be an effective alignment technique, we may want a model to be both \textit{helpful} and \textit{trustworthy}, for example.
In this paper, we push the limits of steering vectors through systematic empirical study, testing \textbf{36 vectors}, across \textbf{two transfer tasks}, and with combinations of up to \textbf{four vectors at once}. These studies contribute the following insights (summarized in Figure \ref{fig:placeholder})\footnote{All experiments and code are available at: 
\url{https://github.com/melaniesubbiah/steering-vectors}
.}:
\begin{itemize}[itemsep=1pt, topsep=1pt]
\item We show that steering effectiveness varies substantially across trait types, with traits that affect global structure and style being most reliably steerable and unconventional or locally-expressed styles being least effective.
\item We demonstrate that steering vectors do not transfer reliably across contexts, with significant drops in trait expression when vectors extracted from style-elicitation prompts are applied to downstream summarization and email-writing tasks.
\item We compare multiple methods for composing steering vectors at inference time, finding that trait expression and output coherence degrade as more vectors are combined, and that no single method eliminates this tradeoff.
\end{itemize}

\section{Related Work}

\paragraph{Activation Steering and Representation Engineering}
Prior work has introduced methods for controlling language models by modifying internal activations rather than updating model weights. \citet{subramani-etal-2022-extracting} showed latent space steering vectors could be used for sentence recovery and unsupervised sentiment transfer. \citet{zou2025representationengineeringtopdownapproach} and \citet{turner2024steeringlanguagemodelsactivation} extend activation engineering to steer model outputs towards specific topics. Other work develops related steering methods for instruction following, safety, and concept control \citep{stolfo2025improving,bhattacharjee2024towards,zhang2025controllinglargelanguagemodels}. Our work builds on this line of research by studying how activation steering behaves when scaled across many stylistic traits, downstream tasks, and vector combinations.

\paragraph{Style and Persona Control in Language Models.}
Recent work applies activation steering specifically toward stylistic traits and personas. Notably, \citet{chen2025personavectorsmonitoringcontrolling} propose vector-based control of character traits. This work builds upon frameworks for translating concepts into linear directions \citep{rimsky-etal-2024-steering, turner2024steeringlanguagemodelsactivation,liu2024incontextvectorsmakingcontext}, and recent approaches leverage linear vectors in representation space for controllable generation \citep{cao2024personalized, dunefsky2025oneshot}. Prior work has primarily focused on a small number of traits or narrowly scoped applications such as safety, hallucinations, sentiment, or persona control. \citet{konen-etal-2024-style} and \citet{zhang2025controllinglargelanguagemodels} extend these ideas to stylistic control, though they evaluate a limited range of style traits. In contrast, we study steering effectiveness systematically across 36 stylistic preferences, two task transfer settings, and four simultaneously applied vectors.

\paragraph{Preference Alignment for Assistive Writing.}
Work on writing-agent alignment has largely approached personalization by inferring natural-language descriptions of user preferences. PRELUDE and CIPHER infer preferences from user edits \citep{gao2024aligning}, while PROSE extends this setup to richer preferences inferred from user samples in PLUME \citep{aroca-ouellette2025aligning}. More general personalization benchmarks such as LaMP evaluate whether language models can adapt to user-specific histories \citep{salemi-etal-2024-lamp}. Related work also studies preference learning from user edits, demonstrations, and content \citep{hewitt2024modeleditingcanonicalexamples,shaikh2025aligning,tan-etal-2025-aligning}. Our work differs from these approaches by studying alignment through steering vectors rather than prompt optimization, editing, or finetuning.

\newcommand{\sep}{\enspace$\cdot$\enspace}

 \begin{table*}[t]
  \centering
  \small
 \begin{tabular}{clp{10.5cm}}
\toprule
\textbf{Task} & \textbf{Source domain} & \textbf{User preferences} \\
\midrule
\multirow{5}{*}{\rotatebox{90}{\textbf{Summarization}}}
  & News (CNN/DM)            & step-by-step structure\sep simile\sep ampersands\sep children's book style \\[4pt]
  & Forum posts (SLF5k)      & header-based structure\sep rhetorical questions\sep ALLCAPS\sep tweet style \\[4pt]
  & Encyclopedia (Wikipedia) & rhyming structure\sep modern slang\sep screenplay style \\[4pt]
  & Paper abstracts (ArXiv)  & Q\&A structure\sep personifications\sep archaic language\sep podcast style \\[4pt]
  & Movie reviews (IMDB)     & stream of consciousness\sep onomatopoeias\sep imagery\sep old timey radio style \\
\midrule
\multirow{4}{*}{\rotatebox{90}{\textbf{Email writing}}}
  & Personal advice (SLF5k)  & intensely emotional\sep alliterations\sep formal tone
\sep second person narrative \\[4pt]
  & Paper reviews (Ampere)   & sharply critical\sep short punchy sentences\sep parenthetical asides\sep assertive \\[4pt]
  & Paper summaries (CC-BY)  & highly inquisitive\sep long flowing sentences\sep emojis\sep conditional expressions \\[4pt]
  & Paper tweets             & sarcastic\sep hyperboles\sep informal tone\sep third person perspective \\
\bottomrule
\end{tabular}
  \caption{Evaluation tasks and user preference sets from PLUME \citep{aroca-ouellette2025aligning}. Each source domain corresponds to a
  distinct set of preferences that are expressible in that domain. }
  \label{tab:plume_dataset}
  \end{table*}

  \begin{figure*}[t]
\begin{center}
\includegraphics[width=\linewidth]{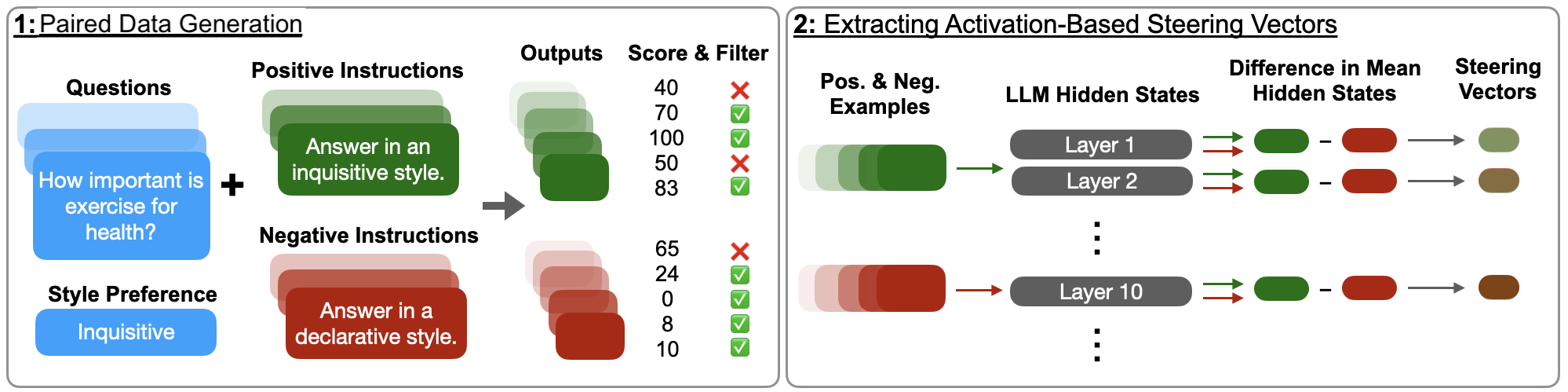}
\end{center}
\caption{To extract a steering vector, we first generate paired positive and negative examples of the stylistic preference, and then extract layer-level differences in activations across these paired examples.} 
\label{fig:methods}
\end{figure*}

\section{Methods}

To consider a setting with many steerable traits, we work on writing personalization. The preferences involved with writing personalization are primarily stylistic rather than safety-related, but we can easily work with a large number of traits that are complementary in this setting.

\subsection{Dataset}

For selecting an appropriate range of traits to test for steering effectiveness and task transfer, we draw from the writing personalization benchmark, PLUME \citep{aroca-ouellette2025aligning}. PLUME was developed in order to address limitations of the assistive writing benchmark PRELUDE \citep{gao2024aligning}. The dataset contains email-writing and summarization tasks, each of which have a set of users with distinct preferences for how each task is completed (shown in Table \ref{tab:plume_dataset}). 

In either task, the model is given a source document to use as content for the summary or email. In addition, it is given preferences meant to control the stylistic traits, examples of which include rhetorical questions, formal tone and all-caps emphasis, or rhyming structure. The source documents and corresponding preferences are sampled from different datasets shown in Table~\ref{tab:plume_dataset}. In the original PLUME task, the model has to iteratively infer and match user preferences from user demos, but we consider one-shot preference alignment using steering vectors for preference traits instead. 
In line with \citet{aroca-ouellette2025aligning} and \citet{gao2024aligning}, we sample 25 examples from each of the source datasets, resulting in 125 tasks for summarization and 100 tasks for email writing\footnote{We release these evaluation sets in the GitHub repository to facilitate future work on these tasks.}. 




\subsection{Steering Vectors}

We extract vectors based on all possible preferences in the PLUME data using the PersonaVectors approach \citep{chen2025personavectorsmonitoringcontrolling}. This method has three main components: (i) paired data generation, (ii) extracting activation-based steering vectors, and (iii) steering output. To consider steering for multiple traits at once, we add (iv) combining multiple traits.
The first two steps are shown in Figure \ref{fig:methods}, and all steps are detailed in this section. 

\paragraph{1. Paired data generation}
We first generate paired instruction data that is designed to elicit and suppress each target trait. Following \citet{chen2025personavectorsmonitoringcontrolling}, we create a set of positive prompts that direct the LLM to generate text tailored towards a target preference and a set of negative prompts that direct it to avoid that preference. To ensure that the vectors are content-agnostic, we pair the instructions with a set of questions across subjects (e.g. "How important is exercise for health?", "What makes a good cup of coffee?"). We use the prompt generation instructions provided by \citet{chen2025personavectorsmonitoringcontrolling} with our own trait definitions shown in Appendix \ref{sec:trait_definitions}.

To generate high-quality questions, we use a powerful LLM. However, the generated responses must be produced by the same model used in the steering pipeline, since steering vectors require direct access to the model's internal activations during generation. We therefore use smaller open-source models for response generation, enabling direct manipulation of activations and extraction of steering vectors. 

\paragraph{2. Extracting activation-based steering vectors} We consider a language model $f_\theta$ with parameters $\theta$ consisting of $L$ transformer layers. 
At each layer $l \in \{1,\dots,L\}$, the hidden representation for token $t$ is denoted 
$h_t^{(l)} \in \mathbb{R}^d$. 
For each stylistic trait $k \in \{1,\dots,K\}$, we precompute an activation-based 
steering vector $v_k^{(l)} \in \mathbb{R}^d$ at selected layers. 
Each steering vector is defined as the difference of mean hidden states between sentences 
with and without the target trait:
\[
v_k^{(l)} = \bar{h}_k^{(l)} - \bar{h}_{\setminus k}^{(l)},
\]
where $\bar{h}_k^{(l)}$ is the mean hidden activation at layer $l$ over a corpus of responses 
exhibiting trait $k$ and across all tokens in the response, and $\bar{h}_{\setminus k}^{(l)}$ is the mean over responses and tokens without the trait.  

\paragraph{3. Steering output}
At inference time, we apply a vector by modifying the residual stream at a specific layer.
$$\tilde{h}_t^{(l)}=h_t^{(l)} + \alpha v_k^{(l)}$$
Where $l$ and $\alpha$ are tuned parameters.

\paragraph{4. Combining multiple traits}
When multiple traits are selected (e.g. both "childlike" and "simile-usage"), we consider different methods of combining steering vectors (full details of the implementation of each method shown in Appendix \ref{sec:composition_methods}):
\begin{itemize}[itemsep=1pt, topsep=1pt]
    \item \textbf{Orthogonalized} – We orthogonalize the vectors, multiply them by their individual alphas, and sum them. 
\item \textbf{Different layers} – We apply each vector at a different layer multiplied by its alpha.
    \item \textbf{Tuned mean} – We multiply each vector by its alpha and take the mean.
\item \textbf{Unit norm} – We take the mean of the unit norms of the vectors and multiply by an alpha.

\end{itemize}


\section{Experimental Setup}
\paragraph{Models} 
To generate the questions and positive and negative stylistic instructions, we use Claude-3.7-Sonnet as a powerful LLM which can handle the nuance of the task. All generated prompts and questions are included in the GitHub repository or can be found in \citet{chen2025personavectorsmonitoringcontrolling}. For each trait, we generate 125 positive and negative questions. 
To generate responses for steering evaluation, we use Qwen2.5-7B-Instruct and Llama3.1-8B-Instruct. Using two similarly sized models allows us to test whether observed steering behaviors and failure modes are consistent across architectures.  
For each model, we generate 100 responses to each question, resulting in 5000 total responses per preference before filtering. We then use the trait expression judge (described below) to verify and filter the generated responses based on trait-expression thresholds, requiring expression greater than 50 for positive examples and lower than 50 for negative examples. We use two 40GB A100 GPUs.
\vspace{-.2cm}
\paragraph{Tasks} For testing a variety of aspects of steering vectors, we use two different task settings:
\begin{itemize}[itemsep=1pt, topsep=1pt]
    \item \textbf{Extraction} tasks – We use the prompts that Claude-3.7-Sonnet generated for each trait to elicit that behavior. These prompts are generally short questions or tasks, such as "How can someone improve their public speaking skills?" or "Describe the benefits of regular exercise." These are the tasks we use to tune steering vector coefficients and test trait expression because these are the prompts used to extract the steering vectors. 
    \item \textbf{PLUME} tasks – We use the PLUME prompts for preference-aligned tasks, which ask the model to write an \textit{email} 
    or a \textit{summary} 
    of some content given in the prompt task. These are the tasks we use to test task transfer and multi-trait expression because this task is designed for preference-aligned generations with sets of four preferences that are complementary \citep{aroca-ouellette2025aligning}.
\end{itemize}
\vspace{-.2cm}
\paragraph{Evaluation}
We use several metrics to assess outputs for trait expression and coherence. These metrics are based on LLM-as-a-judge scores, which have been shown to acheive high agreement with human preferences \citep{zheng2023judging} and have been established by other research. Prompts for the LLM-as-a-judge methods are included in the GitHub repository and taken from \citet{aroca-ouellette2025aligning} and \citet{chen2025personavectorsmonitoringcontrolling}. 

\begin{itemize}[itemsep=1pt, topsep=1pt]
    \item \textbf{PLUME trait expression} – This metric is computed using GPT-4o as the LLM judge, $J$, to assess how closely the generations align with the ground-truth user preferences. Let $J(y, p)$ indicate GPT-4o's judgment of how well a generation $y$ adheres to a preference $p$ (on a 5-point Likert scale from -2 to 2). The Per Preference-Component Match (PPCM) then averages these scores across the ground-truth preferences $\mathcal{P}$:
\[
\text{PPCM}(y, \mathcal{P}) = \frac{\sum_{p \in \mathcal{P}}J(y, p)}{|\mathcal{P}|}
\]
\citet{aroca-ouellette2025aligning} designed this metric for PLUME and verified it has good correlation with human judgments of generation quality and preference adherence. 
\item \textbf{Extraction trait expression} – This metric prompts GPT-4.1-mini to score trait expression on a scale of 0 to 100. \citet{chen2025personavectorsmonitoringcontrolling} designed this evaluation for PersonaVectors and found it has good correlation with human judgments for the traits they assessed: \textit{evil}, \textit{sycophancy}, and \textit{hallucination}. This judge is also used in the paired data generation step of steering vector extraction.
\item \textbf{Coherence score} - This metric prompts GPT-4.1-mini to score output coherence on a scale of 0 to 100. \citet{chen2025personavectorsmonitoringcontrolling} and \citet{Betley_2026} use this evaluation for alignment work. They generally use a threshold of 50 for coherence. We use a stricter threshold of 75 to identify outputs that are unlikely to include repetition or incoherence. 
\end{itemize}
\begin{figure*}[t]
    \centering
    
    \includegraphics[width=\textwidth]{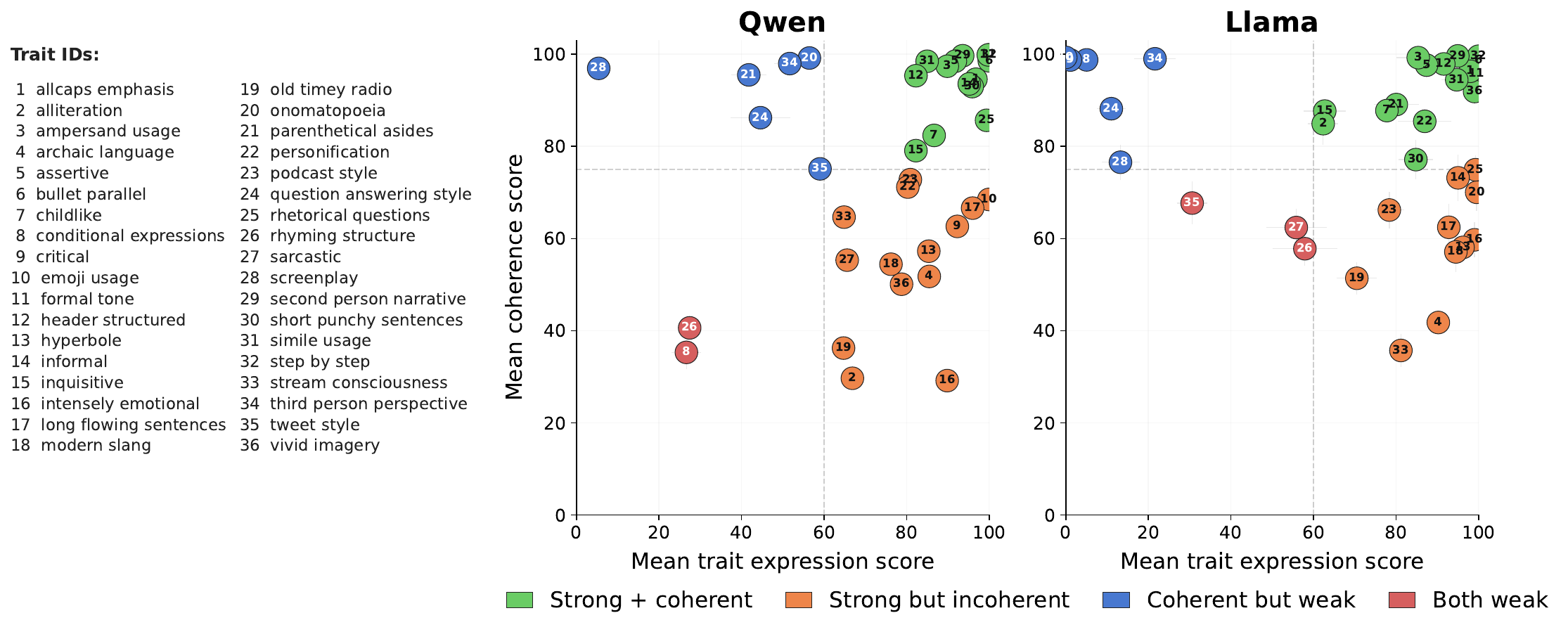}
    
    \vspace{0.6em}
    
    \includegraphics[width=0.75\textwidth]{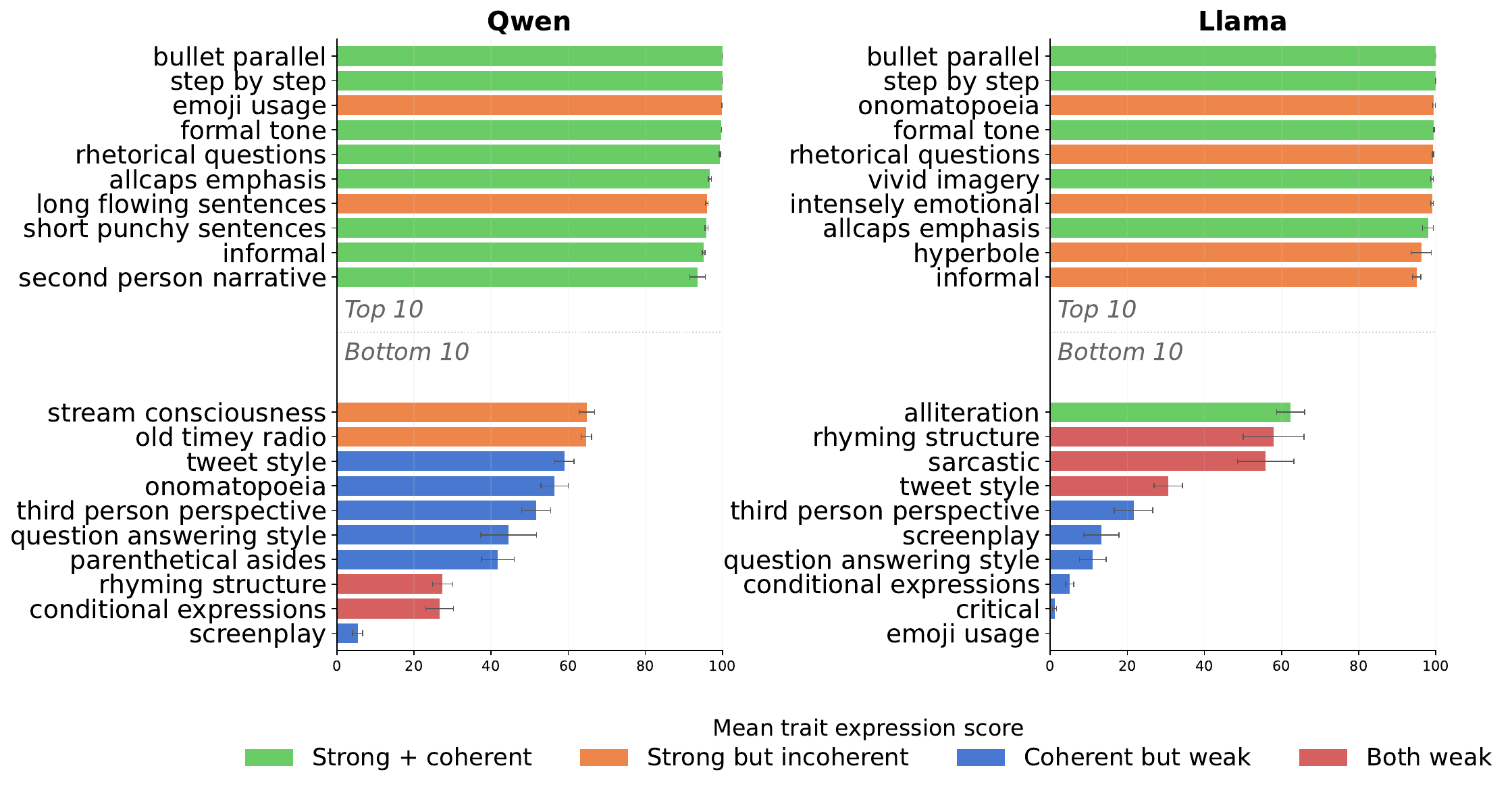}
    
    \vspace{-0.2em}
    
    \caption{
Steering effectiveness varies substantially across stylistic traits and models.
Top: Trait-expression score versus coherence score for each trait.
Bottom: Traits ranked by mean expression score, with colors indicating coherence outcomes.
While many traits remain both expressive and coherent under steering, others exhibit weak controllability or incoherence.
}
    
    \label{fig:steering_effectiveness}
\end{figure*}

\subsection{Alpha and Layer Tuning}
\label{sec:tuning}
 Steering vectors can be applied at any layer in the model and with varying degrees of strength. Since steering vectors directy affect the model's residual stream, interfering too much with the activations (with a high $\alpha$) can lead to incoherence in the output. There is therefore an explicit tradeoff between coherence and trait expression. As a result, we need to tune which layer and strength to use for each vector to ensure optimal performance. 

 We identify candidate layers based on prior work (layers 16 and 20 for the models we use) and conduct an initial coarse search over integer coefficients in [1, 5] at each layer. For each layer-coefficient combination, we test responses to five randomly sampled questions from the question set generated for paired data generation. We select the layer and coefficient that maximize average trait expression while maintaining an average coherence score $\geq 75$; this combination serves as our anchor. We then perform a fine-grained sweep around the anchor coefficient at the anchor layer, testing values at increments of 0.1 over the range [anchor - 0.5, anchor + 0.5], again selecting the configuration with the highest trait expression subject to coherence $\geq 75$. At any stage, the search terminates early if a combination achieves trait expression $\geq 90$ with coherence $\geq 75$.

\section{Results}
We test a variety of properties of how flexibly steering vectors can be applied, considering type of trait, task transfer, combination method for multiple vectors, and number of vectors.

\subsection{Which types of traits can be steered?}

In Figure \ref{fig:steering_effectiveness}, we show how effectively each trait can be steered for each model using the best setting from layer and coefficient tuning. In this plot, we use the same trait expression judge and extraction tasks used for tuning settings, demonstrating the best case scenario for trait expression. For both models, we see there are steering vectors that result in minimal trait expression and/or incoherent output. Qwen has more incoherent outputs, while Llama has more unexpressed traits.

In terms of individual traits, there are some steering vectors that are in the top ten most effective across both models: \textit{bullets}, \textit{step-by-step}, \textit{formal tone}, \textit{rhetorical questions}, \textit{all-caps emphasis}, and \textit{informal tone}. These traits seem to follow common overall structural or stylistic tendencies the models have. 
In contrast, some steering vectors consistently appear in the bottom ten least effective across both models: \textit{tweet style}, \textit{third person perspective}, \textit{question answering style}, \textit{rhyming structure}, \textit{conditional expressions}, and \textit{screenplay}. These traits use more unconventional rhetorical devices 
that do not affect all parts of the output equally, like \textit{rhyming structure} that matters most at the end of each line. Lastly, there are some traits that work very well in one model and not the other, such as \textit{emoji usage} for Qwen and \textit{onomatopoeia} for Llama. Overall, these results show that steering effectiveness depends on context and trait. Many traits cannot be steered reliably without introducing incoherence in outputs. Finally, traits that follow common LLM output styles and affect the structure or style as a whole tend to be the most effective.

\subsection{Does steering transfer across tasks?}
\begin{figure*}[t]
    \centering
    \includegraphics[width=\textwidth]{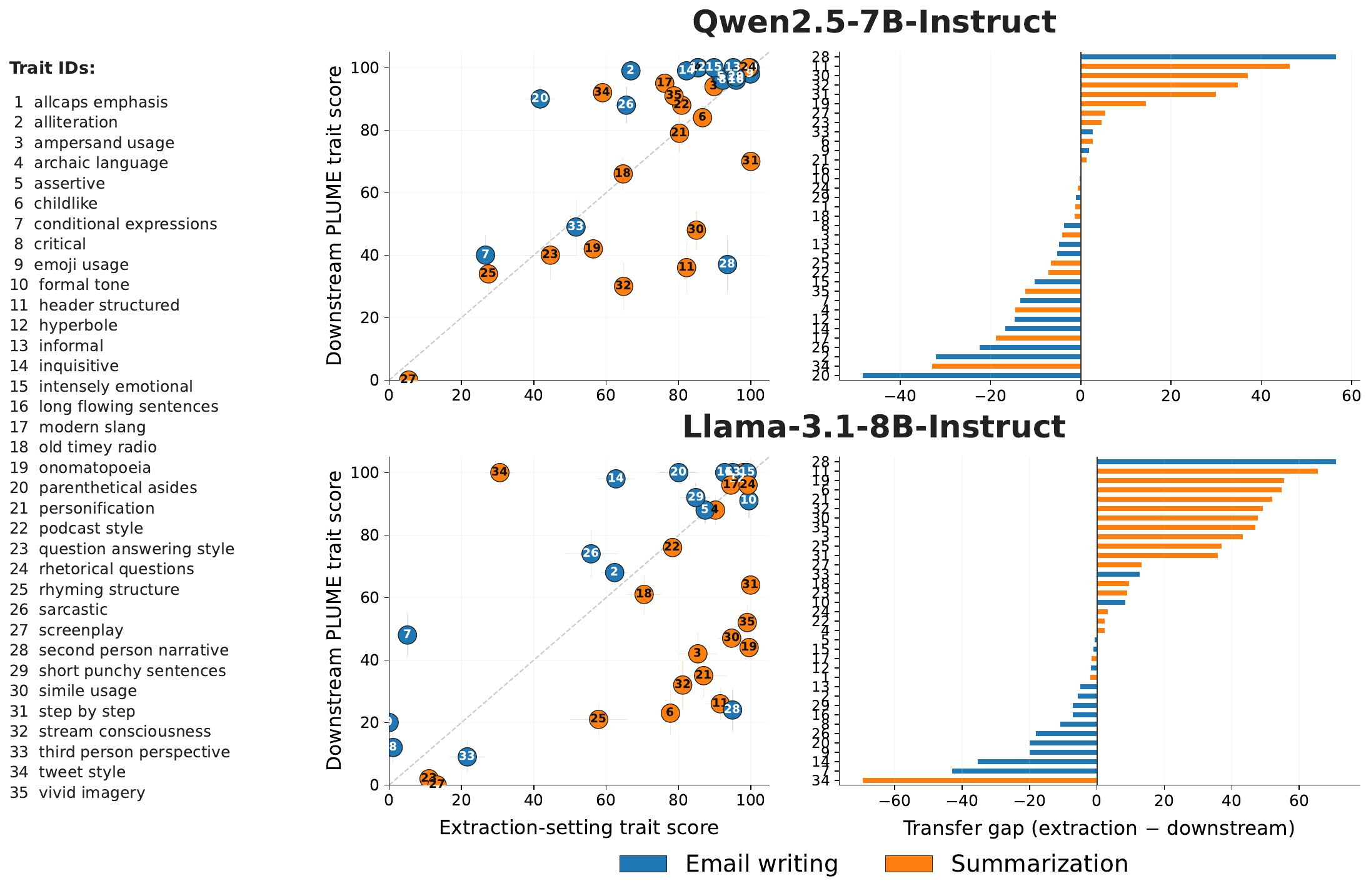}
    \caption{
    Steering transfer from the extraction setting to downstream PLUME tasks. 
    Left: each trait is plotted by its extraction-setting trait score and downstream PLUME trait score; points on the diagonal indicate perfect transfer. 
    Right: transfer gap, defined as extraction score minus downstream score, for each trait. 
    Positive gaps indicate stronger expression in the extraction setting, while negative gaps indicate stronger expression downstream. 
    Email-writing traits are shown in blue and summarization traits in orange.
    }
    \label{fig:steering_transfer}
\end{figure*}
In Figure \ref{fig:steering_transfer}, we investigate whether steering effectiveness in the extraction setting transfers well to the  downstream PLUME tasks: summarization and email writing.
If steering transfers perfectly, we would see all traits aligned along the middle diagonal in the first plot and relatively short bars across all traits in the second plot. We  see that instead, 
there is significant change in steering effectiveness between the extraction and PLUME tasks, demonstrating that steering depends significantly on the context in which it is applied and vectors may be capturing task-dependent aspects of a trait. We also notice that summarization has more traits that degrade significantly in performance. Summarization is typically a more impersonal task, so it may be more unnatural 
for the models to inject particular style into summaries. For both models, the \textit{second person narrative} and \textit{header structured} traits transfer poorly, and for both models, the \textit{tweet style} trait increases significantly in expression. 
Overall, these results indicate steering performance is highly dependent on the task and judge used for vector extraction and coefficient tuning. Performance can then degrade substantially when the resulting vectors are applied in downstream settings.

\subsection{What is the best method to apply multiple steering vectors simultaneously?}

\begin{table*}[t]
\centering  \small
\begin{tabular}{ll|rrrr|r}
\toprule
&\textbf{Method} & \multicolumn{4}{c|}{\textbf{Trait Expression}} & \textbf{Coherence}\\
 &  & Trait 1 & Trait 2 & Average & $\Delta$ vs baseline &  \\
\midrule\midrule
\multirow{5}{*}{\rotatebox{90}{\textbf{Qwen}}} & Orthogonalized & $45.2{\,{\scriptstyle\pm 4.2}}$ & $48.3{\,{\scriptstyle\pm 4.4}}$ & $46.8{\,{\scriptstyle\pm 3.7}}$ & $-28.5{\,{\scriptstyle\pm 2.7}}$ & $36.8{\,{\scriptstyle\pm 3.3}}$ \\
 & Different layers & $51.3{\,{\scriptstyle\pm 3.9}}$ & $48.5{\,{\scriptstyle\pm 3.9}}$ & $49.9{\,{\scriptstyle\pm 3.2}}$ & $-26.6{\,{\scriptstyle\pm 2.5}}$ & $40.5{\,{\scriptstyle\pm 3.2}}$ \\
 & Tuned mean & $43.2{\,{\scriptstyle\pm 4.4}}$ & $47.6{\,{\scriptstyle\pm 4.6}}$ & $45.4{\,{\scriptstyle\pm 3.6}}$ & $-29.9{\,{\scriptstyle\pm 2.8}}$ & $47.9{\,{\scriptstyle\pm 4.2}}$ \\
 & Unit norm (layer=16, $\alpha$=30) & $39.4{\,{\scriptstyle\pm 5.4}}$ & $33.5{\,{\scriptstyle\pm 5.2}}$ & $36.5{\,{\scriptstyle\pm 3.7}}$ & $-40.1{\,{\scriptstyle\pm 3.5}}$ & $97.0{\,{\scriptstyle\pm 1.2}}$ \\
 & Unit norm (layer=20, $\alpha$=58) & $63.0{\,{\scriptstyle\pm 5.5}}$ & $49.7{\,{\scriptstyle\pm 5.7}}$ & $56.4{\,{\scriptstyle\pm 4.1}}$ & $-20.1{\,{\scriptstyle\pm 3.3}}$ & $75.9{\,{\scriptstyle\pm 3.5}}$ \\\midrule
\multirow{5}{*}{\rotatebox{90}{\textbf{Llama}}} & Orthogonalized & $63.9{\,{\scriptstyle\pm 4.4}}$ & $49.0{\,{\scriptstyle\pm 5.1}}$ & $56.5{\,{\scriptstyle\pm 3.7}}$ & $-15.7{\,{\scriptstyle\pm 2.9}}$ & $56.0{\,{\scriptstyle\pm 4.0}}$ \\
 & Different layers & $55.2{\,{\scriptstyle\pm 4.0}}$ & $50.7{\,{\scriptstyle\pm 4.1}}$ & $53.0{\,{\scriptstyle\pm 3.1}}$ & $-18.9{\,{\scriptstyle\pm 2.4}}$ & $47.7{\,{\scriptstyle\pm 3.4}}$ \\
 & Tuned mean & $61.9{\,{\scriptstyle\pm 4.6}}$ & $42.8{\,{\scriptstyle\pm 4.9}}$ & $52.3{\,{\scriptstyle\pm 3.6}}$ & $-19.9{\,{\scriptstyle\pm 2.9}}$ & $65.0{\,{\scriptstyle\pm 4.1}}$ \\
 & Unit norm (layer=16, $\alpha$=4) & $51.0{\,{\scriptstyle\pm 5.9}}$ & $41.4{\,{\scriptstyle\pm 5.8}}$ & $46.2{\,{\scriptstyle\pm 4.2}}$ & $-25.7{\,{\scriptstyle\pm 3.5}}$ & $94.5{\,{\scriptstyle\pm 1.0}}$ \\
 & Unit norm (layer=20, $\alpha$=8) & $55.8{\,{\scriptstyle\pm 6.0}}$ & $42.6{\,{\scriptstyle\pm 5.7}}$ & $49.2{\,{\scriptstyle\pm 4.7}}$ & $-22.7{\,{\scriptstyle\pm 3.3}}$ & $87.6{\,{\scriptstyle\pm 2.5}}$ \\
\bottomrule
\end{tabular}
\caption{Comparing the effectiveness of different methods for combining multiple steering vectors at inference time (reported as mean $\pm$ SEM). Trait and coherence scores are 0--100. $\Delta$ vs baseline is the signed difference from the single-trait score.}
\label{tab:q3_method_comparison}
\end{table*}

In Table \ref{tab:q3_method_comparison}, we report the results of trying to apply two steering vectors simultaneously at inference time. We evaluate these combinations using the same setting and judge used during vector extraction to isolate combination method from task transfer.
We can see that for all methods, trait expression drops significantly (at least 15\% reduction). Many outputs are also somewhat incoherent (considered as anything with a coherence score of $<75\%$). In many cases, one trait also dominates over another to some degree. This effect seems more prominent in the unit norm methods. In these methods, traits are balanced equally rather than allowing re-balancing effects from different coefficients.
However, we see that for both models, the unit norm methods produce significantly more coherent outputs. This suggests that placing steering vectors on different layers can amplify the effects of each trait simultaneously, leading to incoherent outputs. In contrast, applying multiple vectors at the same layer may allow the model to balance competing effects.
The only method which reduces trait expression only moderately while maintaining output coherence across both models is the unit norm approach applied at layer 20. However, for both models, the trait expression for the two traits is fairly imbalanced in this approach, demonstrating that there are different tradeoffs of different methods. 

\subsection{What is the effect of increasing the number of combined steering vectors?}

\begin{figure*}[t]
    \centering
    \includegraphics[width=\textwidth]{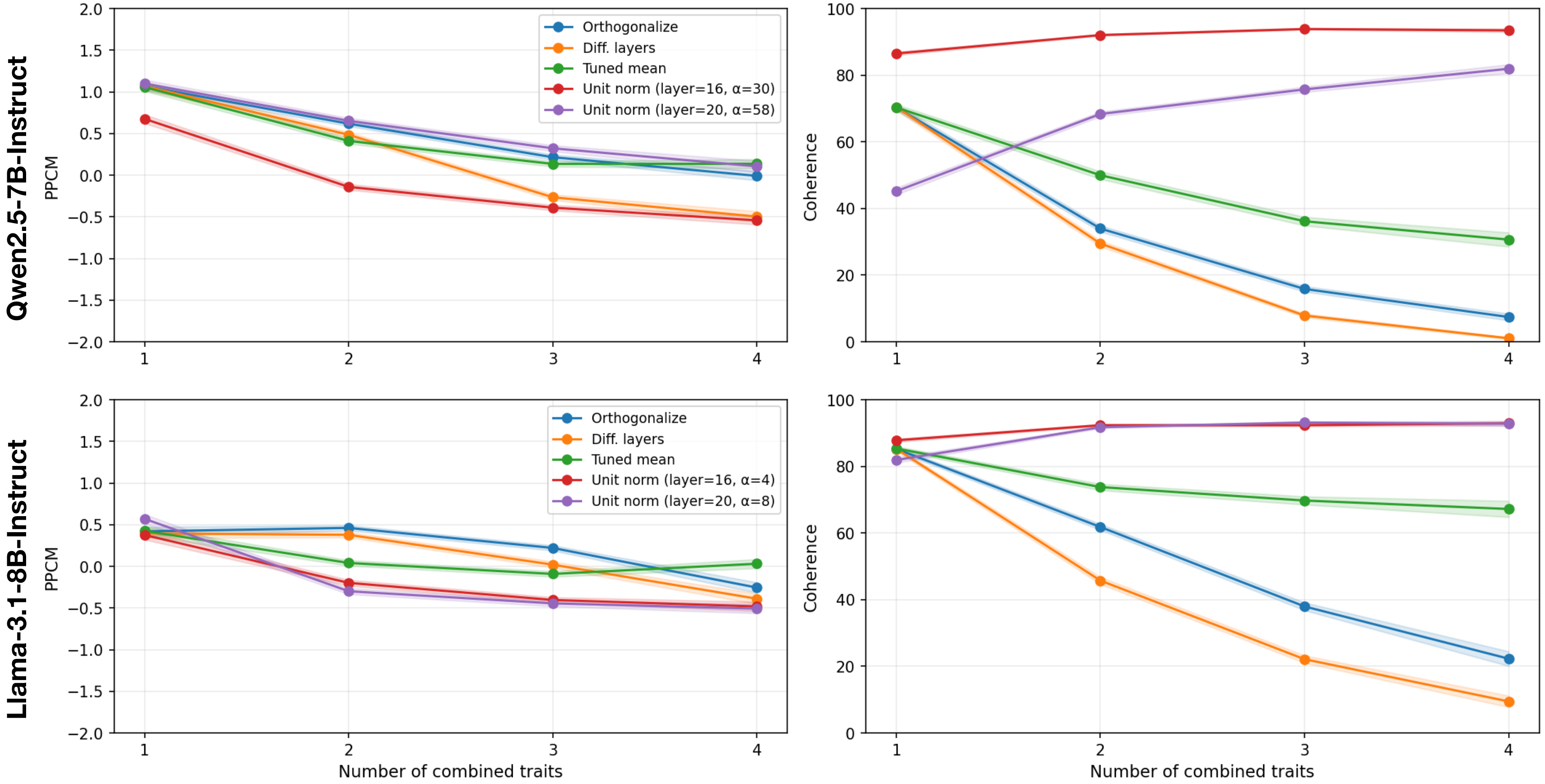}
    \caption{We show how steering effectiveness changes as more traits are steered simultaneously. The left plot shows the PPCM score on PLUME tasks. The right pot shows the coherence score for outputs. We compare the different trait combination methods.}
    \label{fig:steering_scale}
\end{figure*}

We expand on the previous question to also see how these combination methods scale to more steering vectors on the PLUME tasks. We show plots for trait expression scores on the left in Figure \ref{fig:steering_scale} and coherence scores on the right. We can see that the unit norm methods remain substantially more coherent across more traits, while the other methods cross into incoherence. The trait expression decreases across all methods as more traits are added. Most methods have negative or no trait expression once their are four traits. For Qwen, the unit norm method at layer 20 does the best job of balancing trait expression and coherence, but for Llama, the tuned mean method seems to be the best. 
Given that the coherence score actually increases as more traits are added for the unit norm methods, it seems that the alpha value likely needs to be increased as more traits are added to balance trait expression and coherence. 
While tuning this parameter could improve coherence, it exposes yet another parameter that requires tuning to apply steering vectors.

\section{Conclusion}
Through experimenting with a range of steering vectors across different tasks and combination settings, we uncover many limitations in their flexible application. We see that some traits are not effectively expressed through steering vectors and transferring steering across tasks depends on how natural steering is in the target task. We find that taking a mean of unit normalised vectors and applying an alpha seems to be the most effective method of combining vectors at inference time, but steering effectiveness and coherence still decreases as more traits are combined. This alpha may additionally require re-tuning each time the number of traits is increased. Collectively, these results reveal that relying on steering vectors for interpretability and alignment will pose limits on generality. 

In addition to issues for controllable generation, the task transfer results indicate there can be generalization issues for alignment research relying on steering vectors to monitor or indicate trait expression in training data or inputs. This may be an effective approach for data that is similar enough to the vector extraction setting but will perform unreliably on data from other downstream settings.


\section*{Limitations}

There are many parameters requiring tuning in the application of steering vectors. We cannot exhaustively tune each of these given time and resource limits around attaching different layer hooks for each setting and using LLM-as-a-judge scores for evaluation. Additionally, selecting the perfect coefficients for four steering vectors applied simultaneously is a complex optimization problem. As a result, there are places where coefficients or layers could be tuned further. Despite this limitation, we see consistent patterns in our results that lead to our conclusions. 

Since we need to modify and extract the residual stream from the model, we are limited to working with open-source models. For the same reasons around resource constraints described above, we work with two commonly used models, but we are therefore not able to conclude whether the limitations found in this study apply to much larger models as well. However, the conclusions in this study can still provide useful pointers to future work around what is most likely to work or what to look out for in applications with larger models.

Finally, the evaluation in this work relies on LLM-as-a-judge metrics. Given the extensive experimentation with different settings required, it is not feasible to use human-in-the-loop evaluation. The metrics we use were designed by prior work and validated against human judgments. Because the focus of our study is comparing the tradeoffs and effects of different decisions in the application of steering vectors (including choice of task judge), using LLM-based scoring is an effective method of demonstrating these tradeoffs at a coarse level.



\bibliography{custom}

\appendix

\section{Trait Definitions}
\label{sec:trait_definitions}

\begin{itemize}[nosep]
    \item \textbf{write assertively}: use assertive language in the response
    \item \textbf{adopt a step-by-step structure}: break the response into individual steps
\item    \textbf{include rhetorical questions}: ask rhetorical questions in the response
\item    \textbf{use semicolons (;) when possible}: use semi-colons when possible in the response
    \item \textbf{include personifications}: include personifications of the details described in the response
\item    \textbf{write in the style of old timey radio}: write the response as though it is part of the transcript of an old timey radio broadcast
\item    \textbf{include a simile}: use similes in the response
\item    \textbf{use imagery}: include vivid imagery to describe aspects of the response
\item    \textbf{include modern slang}: incorporate modern slang in the response
\item    \textbf{use ampersands (\&) instead of "and"s}: use ampersands in the response
\item    \textbf{use ALLCAPS to emphasize certain words}: capitalize all the letters in some words to emphasize their importance in the response
\item    \textbf{use archaic language}: use old-fashioned language in the response as though writing from a much earlier century
\item    \textbf{write in the style of a podcast}: write the response as though it is part of the transcript of a podcast episode
\item    \textbf{sign off the email using an epithet}: conclude the response by signing off with an epithet, meaning using an adjective or descriptive phrase instead of a name
\item    \textbf{adopt a rhyming structure}: write the response with rhyming lines
\item    \textbf{adopt a question-answering style structure}: write the response as a series of questions and answers
\item    \textbf{include alliterations}: use some alliteration in the response, but do not write entire sentences with alliteration
\item    \textbf{include hyperboles}: use hyperbole, meaning lots of exaggeration, in the response
\item    \textbf{be intensely emotional}: use intensely emotional language in the response
\item    \textbf{be sharply critical}: write the response using critical and hostile language
\item    \textbf{be highly inquisitive}: write the response using an inquisitive and questioning style
\item    \textbf{be blatantly sarcastic}: write the response using lots of clear sarcasm
\item    \textbf{include several long and flowing sentences}: include many long, flowing sentence in the response
\item    \textbf{open the email using a movie reference}: start the response with a reference to a movie
\item    \textbf{use parenthetical asides}: use many parenthetical asides within the response
\item    \textbf{write using conditional expressions}: write the response using conditional expressions (e.g., if/then statements)
\item    \textbf{write using bullet points}: include bullet points in the response
\item    \textbf{use emojis}: use emojis in the response
\item    \textbf{include several short and punchy sentences}: focus the response on using short, impactful sentences to make a statement
\item    \textbf{write using a stream-of-consciousness style}: write using a stream-of-consciousness style with one thought flowing directly into another
\item    \textbf{write in the style of a tweet}: write the response in the style of a tweet
\item    \textbf{include onomatopoeia}: use onomatopoeia in the response, meaning words that sound like what they are referring to
\item    \textbf{write in the style of a children's book}: write the response as though it is a written story for a child
\item    \textbf{write in the style of a screenplay}: write the response as though it is part of a screenplay for a movie
\item    \textbf{adopt a second person narrative}: write the response using second person narration
\item    \textbf{adopt a third person narrative}: write the response using third person narration
\item    \textbf{use a formal tone}: write the response in a formal and professional tone
\item    \textbf{use an informal tone}: write the response in an informal and casual tone
\item    \textbf{adopt a header and sub-header structure}: write the response with header and sub-header structure

\end{itemize}

\section{Composition Methods}
\label{sec:composition_methods}

 Before combining vectors, each trait $k$ is individually tuned to find the best layer $\ell_k^* \in \{16, 20\}$ and coefficient $\alpha_k^* $ using the approach describe in Section \ref{sec:tuning}. Given two traits $k_1$ and $k_2$ with tuned parameters $(\ell_{k_1}^*, \alpha_{k_1}^*)$ and $(\ell_{k_2}^*, \alpha_{k_2}^*)$, we compare four ways to compose their vectors. For clarity, we focus on two traits here, but each method generalizes to more traits by following the same pattern.

   \textbf{Orthogonalized.}
  When both traits share the same tuned layer, to reduce interference between vectors that share a direction in representation space, the second vector is projected to remove its component along the first
  before application:
  \begin{equation}
      \mathbf{v}_{k_2}^{\perp} = \mathbf{v}_{k_2} - \bigl(\mathbf{v}_{k_2} \cdot \hat{\mathbf{v}}_{k_1}\bigr)\,\hat{\mathbf{v}}_{k_1},
  \end{equation}
  where $\hat{\mathbf{v}}_{k_1} = \mathbf{v}_{k_1}/\|\mathbf{v}_{k_1}\|$. Then $\alpha_{k_1}^*\,\mathbf{v}_{k_1}$ and $\alpha_{k_2}^*\,\mathbf{v}_{k_2}^{\perp}$ are
 summed and applied at $\ell^*$. When traits have different tuned layers, this reduces to the different-layers method.
  
  \textbf{Different layers.}
  Each vector is applied independently at its own tuned layer:
  \begin{equation}
      \tilde{h}_t^{(\ell_k^*)} = h_t^{(\ell_k^*)} + \alpha_k^*\,\mathbf{v}_{k}, \quad k \in \{k_1, k_2\}.
  \end{equation}
  When both traits share the same tuned layer ($\ell_{k_1}^* = \ell_{k-2}^*$), $\mathbf{v}_{k_2}$ is shifted down one layer and still applied with $\alpha_{k_2}^*$.
  
  \textbf{Tuned mean.}
  When both traits share the same tuned layer, their scaled vectors are averaged and applied as a single intervention:
  \begin{equation} 
      \tilde{h}_t^{(\ell_{k_{1,2}}^*)} = h_t^{(\ell_{k_{1,2}}^*)} + \frac{\alpha_{k_1}^*\,\mathbf{v}_{k_1} + \alpha_{k_2}^*\,\mathbf{v}_{k_2}}{2}.
  \end{equation}
  When traits have different tuned layers, this reduces to the \textit{different layers} method.

  \textbf{Unit norm.}
  Each vector is unit-normalised before combining with a mean, and the result is applied at a fixed layer $\ell$ with a single shared coefficient $\alpha$:
  \begin{equation}
      \tilde{h}_t^{(\ell)} = h_t^{(\ell)} + \alpha \cdot \frac{1}{n}\sum_{j=1}^{n}
  \frac{\mathbf{v}_{k_j}}{\|\mathbf{v}_{k_j}\|}.
  \end{equation}
  This decouples the magnitude of each vector from the trait-specific tuning, making it straightforward to extend to $n > 2$ traits with a single scalar to
  tune. We evaluate $\ell \in \{16, 20\}$. For our experiments, $\alpha$ is fixed per model and layer: for Qwen, $\alpha$ is 30 at layer 16 and 58 at layer 20; for Llama, $\alpha$ is 4 at layer 16 and 8 at layer 20. 
   We set these $\alpha$ values as twice the mean norm of steering vectors at that layer (rounded to the nearest integer), which we found to be a reasonable value in early experiments. Future work could further tune this value.



    
    


\end{document}